\documentclass[conference]{IEEEtran}
\IEEEoverridecommandlockouts
\usepackage{cite}
\usepackage{amsmath,amssymb,amsfonts}
\usepackage{algorithmic}
\usepackage{graphicx}
\usepackage{textcomp}
\usepackage{xcolor}
\graphicspath{{figs/}}
\usepackage{booktabs}
\usepackage{etoolbox}
\usepackage{siunitx}
\usepackage{textcomp}
\usepackage{multirow}
\usepackage{siunitx}

\usepackage[caption=false, font=footnotesize]{subfig}
\usepackage[export]{adjustbox}
\def\BibTeX{{\rm B\kern-.05em{\sc i\kern-.025em b}\kern-.08em
    T\kern-.1667em\lower.7ex\hbox{E}\kern-.125emX}}

\usepackage{pifont}
\newcommand{\cmark}{\ding{51}}%
\newcommand{\xmark}{\ding{55}}%
\begin{document}

\title{Privacy-Preserving Image Classification Using Vision Transformer
}

\author{\IEEEauthorblockN{Zheng Qi, AprilPyone MaungMaung, Yuma Kinoshita and Hitoshi Kiya}
\IEEEauthorblockA{Tokyo Metropolitan University, Asahigaoka, Hino-shi, Tokyo, 191--0065, Japan}
}

\maketitle

\begin{abstract}
In this paper, we propose a privacy-preserving image classification method that is based on the combined use of encrypted images and the vision transformer (ViT). The proposed method
allows us not only to apply images without visual information to ViT models for both training and testing but to also maintain a high classification accuracy. ViT utilizes patch embedding and position embedding for image patches, so this architecture is shown to reduce the influence of block-wise image transformation. In an experiment, the proposed method for privacy-preserving image classification is demonstrated to outperform state-of-the-art methods in terms of classification accuracy and robustness against various attacks. 
\end{abstract}


\section{Introduction}
The spread of deep neural networks (DNNs)~\cite{m1} has greatly contributed to solving complex tasks for many applications, such as computer vision, biomedical systems, and information technology. Recently, it has been very popular for data owners to utilize cloud servers such as Google Cloud and Amazon Web Server to compute and process a large amount of data instead of using local servers. This is because cloud environments provide flexibility and cost-saving computation. However, DNNs have been deployed in security-critical applications, such as facial recognition, biometric authentication, and medical image analysis. Since cloud servers are not trusted in general, data privacy, such as personal information and medical records, may be compromised. Therefore, it is necessary to protect data privacy in cloud environments, so privacy-preserving DNNs have become an urgent challenge~\cite{p2,n10}. Homomorphic encryption methods~\cite{h1,h2,h3,h4,h5,h6,h7,h8} may contribute to such a problem, but the computation and memory costs are high, and it is not easy to apply these methods to DNNs directly.

In this paper, we focus on protecting data privacy by encrypting data before uploading the data to the cloud environment. Various perceptual encryption methods, referred to as learnable encryption, have been proposed so that encrypted images can be directly applied to DNN models. However, conventional methods degrade the classification performance compared with models trained with plain images, and moreover, they are not robust enough against attacks~\cite{m5,n4,n5}. To overcome these problems, we focus on using the vision transformer (ViT)~\cite{m2}, which was proposed to solve image classification tasks. We point out that two unique properties due to two embeddings in the architecture of ViT allow us to reduce the influence of block-wise image encryption methods. We propose a novel privacy-preserving image classification method that uses ViT and a block-wise encryption method based on the properties of ViT.

In an experiment, the proposed method was compared with state-of-the-art methods in terms of classification accuracy and robustness against various attacks to demonstrate its effectiveness.

\section{Related Work}
Privacy-preserving image classification and ViT are briefly summarized here.

\subsection{Privacy-Preserving Image Classification}

To apply images without the visual information of plain images to a DNN model, various perceptual image encryption methods, referred to as learnable image encryption, have been proposed. Tanaka first introduced a block-wise learnable image encryption (LE) method with an adaptation layer that is used prior to the classifier to reduce the influence of image encryption~\cite{n1}. Another encryption method is a pixel-wise encryption (PE) method in which negative-positive transformation and color component shuffling are applied without using any adaptation layer~\cite{m4}. However, both encryption methods are not robust enough against ciphertext-only attacks as reported in~\cite{m7,m5}. To enhance the security of encryption, Tanaka's method was extended by adding a block scrambling step and utilizing different block keys for the pixel encryption operation~\cite{n2} (hereinafter denoted as ELE). However, ELE still has a lower accuracy than that when using plain images, although the security of the encryption is improved.

In contrast, image encoding approaches for privacy-preserving image classification have been proposed to hide visual information, where the encoding approaches use no key~\cite{n9,m7}, but they do not consider the ability to decode images. One method for the encoding approaches utilizes a generative adversarial network (GAN) to generate images without visual information with the help of a pre-trained classification model~\cite{n9}. However, it is not robust enough against attacks as reported in~\cite{n5}. Another method uses a transformation network with U-Net, which is also trained by incorporating it with a pre-trained classifier to protect visual information during inference time~\cite{m7}. This method is robust against various attacks, but it is not useful for training a model. Accordingly, we propose a novel privacy-preserving classification method to improve these issues that the conventional methods have. A privacy-preserving method with ViT was proposed in \cite{m10}, but its classification accuracy is lower than the proposed method, although compressible encrypted images can be applied to it.

\begin{figure}[htbp]
	\centerline{\includegraphics[width=8.5cm]{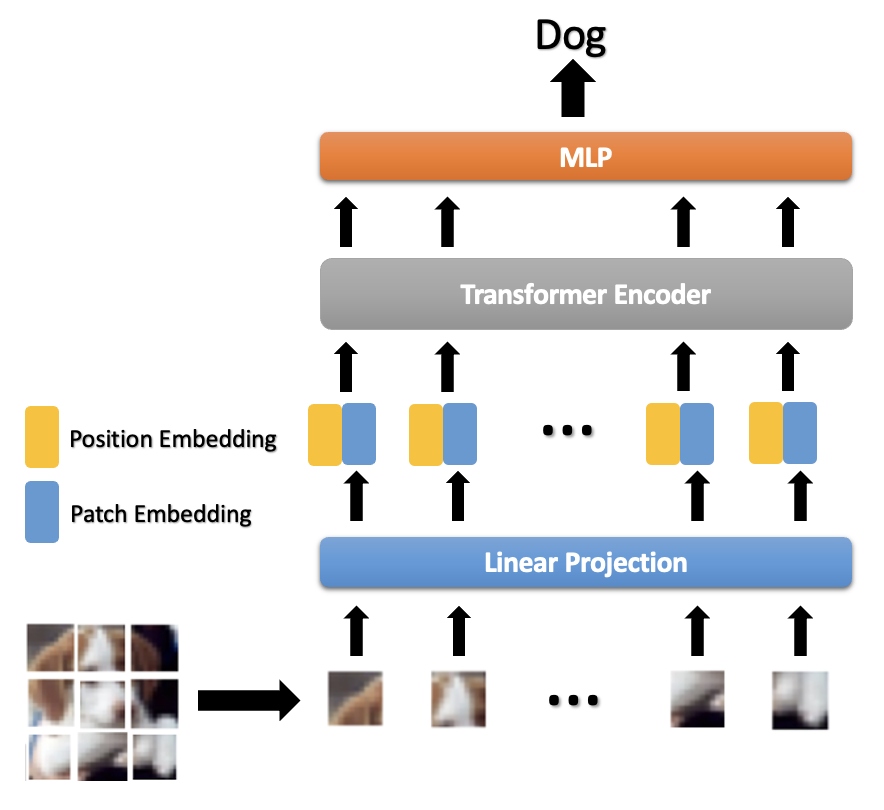}}
	\caption{Architecture of vision transformer.\label{fig:vit}}
\end{figure}

\subsection{Vision Transformer}
The transformer architecture has been widely used in natural language processing (NLP) tasks~\cite{n3}. Vision transformer (ViT) has also produced excellent results compared with state-of-the-art convolutional networks. Figure~\ref{fig:vit} illustrates the architecture of ViT. ViT utilizes patch embedding and position embedding, which are added together, and the resulting embedding is used as an input to the transformer encoder. 

In this paper, we point out that ViT has two unique properties due to these embeddings, so the use of ViT can be expected to maintain a high classification accuracy even under the use of a block-wise encryption method.

\section{Proposed Image Classifier with ViT}
A novel privacy-preserving image classifier is proposed here.

\begin{figure}[htbp]
	\centerline{\includegraphics[width=8.5cm]{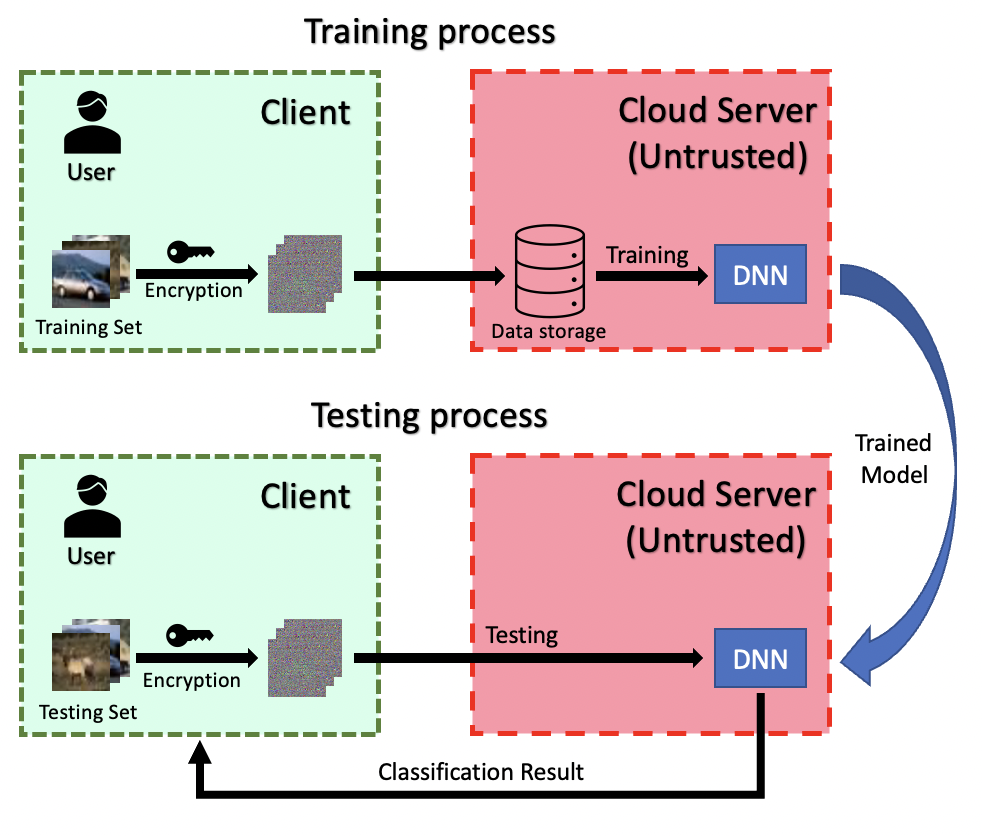}}
	\caption{Framework of proposed method.\label{fig:proposed}}
\end{figure}

\subsection{Overview}
Figure~\ref{fig:proposed} shows the framework of the proposed method. Cloud servers are not trusted in general, because the providers may illegally access the data uploaded by a user or accidentally leak the data.

A user encrypts training images by using a secret key and then sends the encrypted images with labels to a provider. The provider trains a model by using the encrypted images. A test image is also encrypted by using the same key as that used for encrypting the training images, and the encrypted image is sent to the provider. A label is estimated with the trained model, and the estimated label is returned to the user. 

Note that the provider has no both visual information of plain images and the key. Privacy-preserving image classifiers are required to satisfy the following requirements:

\begin{itemize}
  \item Having a high classification accuracy and
  \item Being robust enough against various attacks.
\end{itemize}

In this paper, to satisfy the above requirements, the combined use of ViT and a block-wise image encryption method is proposed for privacy-preserving image classification.

\begin{figure}[htbp]
	
	\begin{minipage}[b]{.48\linewidth}
		\centering
		\centerline{\includegraphics[width=3.5cm]{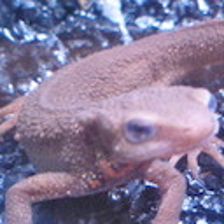}}
		\centerline{(a) Plain}\medskip
	\end{minipage}
	\hfill
	\begin{minipage}[b]{0.48\linewidth}
		\centering
		\centerline{\includegraphics[width=3.5cm]{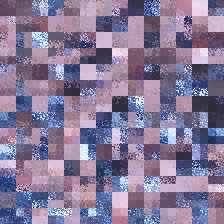}}
		\centerline{(b) Encrypted}\medskip
	\end{minipage}
	\caption{Example of plain image and encrypted image.\label{fig:example}}

\end{figure}

\subsection{Proposed  Encryption Method with ViT}
In this section,we propose a novel encryption method with ViT, where the patch-size of ViT is $ M\times M$.

The procedure of the proposed encryption is summarized as follow:
\begin{enumerate}
	\item Divide an image into blocks with a size of $M \times M$ as
		$B=\{B_1,\ldots,B_i,\ldots,B_N\}, i \in \{ 1,\ldots,N \}$, where $N$ is the number of blocks, and $B_i$ is a divided block.
	\item Generate a random integer $ K_{1} $ as\\
	$ K_{1}=[\alpha_{1}, \ldots, \alpha_{i}, \ldots, \alpha_{i'}, \ldots, \alpha_{N}]$,
	where $\alpha_{i} \in \{1,\ldots,N\}$ and $\alpha_{i} \neq \alpha_{i'}$ if $i \neq i'$.
	\item Permute the blocks with $K_{1}$ such that $B'_{i}=B_{\alpha_{i}}$ and permuted blocks are given by $B'=\{B'_{1}, \ldots, B'_{i}, \ldots, B'_{N}\}$.
	\item Divide each block into four sub-blocks with a size of $S=\dfrac{M}{2} \times \dfrac{M}{2}$  as
    $b=\{b_{1}, \ldots, b_{j}, \ldots, b_{4N}\}, j \in \{1, \ldots, 4N\}.$
	\item Flatten each sub-block $b_{j} $ as a vector as\\
	$v_{j}=[v_{j}(1),\ldots,v_{j}(k),\ldots,v_{j}(S)], k \in \{1, \ldots, S\}$.
	\item Generate a random integer $K_{2}$ as \\ $K_{2}=[\beta_1, \ldots, \beta_k, \ldots, \beta_{k'}, \ldots, \beta_{S}]$, where $\beta_{k} \in \{1,\ldots,S\}$ and $\beta_k \neq \beta_{k'}$ if $k \neq k'$.
	\item Shuffle each vector $v_{j}$ with key $ K_{2} $ such that $v'_{j}(k)=v_{j}(\beta_k)$ and a shuffled vector is given by $v'_{j}=[v'_{j}(1),\ldots,v'_{j}(S)]$.
	\item Concatenate all the sub-blocks to generate an encrypted image.
\end{enumerate}

Figure~\ref{fig:example} is an example of encrypted images. ViT utilizes patch embedding and position embedding (see Fig.~\ref{fig:vit}) and has the following two properties.

\begin{enumerate}
  \item Patch-order invariance of transformer encoder: the output of the transformer encoder corresponding to an input patch is independent of the order of input patches.
  \item Ability to adapt to pixel order by patch embedding: patch embedding can be
adapted to pixel shuffling because pixel shuffling can be expressed as an invertible linear transformation, and patch embedding is done by a learnable linear transformation.
\end{enumerate}

Property 1 allows us to reduce the influence of block scrambling when the patch size of ViT is equal to the block size of image encryption, and property 2 allows us to reduce the influence of pixel shuffling. Accordingly, ViT can be expected to maintain a high accuracy even under the use of encrypted images.

\subsection{Security Evaluation}
The security of the proposed method is evaluated in terms of the key space and the robustness against various attacks, where it is assumed that the encryption algorithm is disclosed except the key, and attackers can obtain only encrypted images.

\subsubsection{Key Space}

The key space of the proposed method should be evaluated. The key space describes a set of all possible keys in an encryption algorithm.

For the case where an image is divided into blocks with a size of $M \times M$, and the number of blocks in an image is $ N $, the key space of the proposed algorithm is given as below.

\begin{equation}
	S_{proposed} = N! \cdot \left(  \frac{M}{2} \right)^2! \label{5}
\end{equation}

For example, when the patch-size of ViT is $ 16\times 16$ and the image size is $ 224\times 224 $, the number of blocks is  $N=196 $ blocks. Hence, the key space is given as follows.

\begin{equation}
	S_{proposed} = 196! \cdot 64! \label{6}=2^{1511}
\end{equation}

The use of a large key space enhances robustness against various attacks.

\subsubsection{Robustness Against Attack Methods}
A lot of attack methods have been studied to restore the visual information of plain images from encrypted ones. Accordingly, the proposed method has to be evaluated in terms of robustness against these attacks. In this paper, three state-of-the-art attack methods, the feature reconstruction attack (FR-attack)~\cite{m5}, the generative adversarial network based attack (GAN-attack)~\cite{n4}, and the inverse transformation network attack (ITN-attack)~\cite{n5}, are used for evaluation, where all of the attacks are ciphertext-only attacks. The FR-attack attempts to restore visual information by using the edge information of images. The GAN-attack and ITN-attack are DNN-based attacks. In an experiment, the proposed method will be shown to be robust against these attacks.

\robustify\bfseries
\sisetup{table-parse-only,detect-weight=true,detect-inline-weight=text,round-mode=places,round-precision=2}
\begin{table}[htbp]
	\caption{Accuracy (\SI{}{\percent}) and security of image classification using various image encryption methods and networks\label{tab:results}}
	\centering
	\begin{tabular}{llSSc}
		\toprule
		\multirow{2}{*}{\bfseries Encryption} & \multirow{2}{*}{\bfseries Network} & \multicolumn{2}{c}{\bfseries Accuracy}& \multirow{2}{*}{\bfseries Security}\\
		& & {\bfseries CIFAR-10} & {\bfseries CIFAR-100}\\
		\midrule
		{LE$^\ast$} & ShakeDrop & 94.49 & 75.48 & {\xmark} \\
		{EtC} & ShakeDrop & 85.94 & 61.90 &{\cmark}\\
		{ELE$^\ast$} & ShakeDrop & 83.06 & 62.97 &{\cmark}\\
		{PE} & ResNet-18 & 92.03 & {--} & {\xmark}\\
		{Proposed} & ViT-B\_16 & \bfseries \num{96.64} & \bfseries \num{84.42} &{\cmark}\\
		\midrule
		Plain & ResNet-18 & 95.53 & {--} & {--} \\
		Plain & ShakeDrop & 96.70 & 83.59 & {--} \\
		Plain & ViT-B\_16 & 99.11 & 92.48 &{--} \\
		\bottomrule
		\multicolumn{4}{l}{$^{\ast}$ An additional adaptation network is applied.}\\
		\multicolumn{5}{l}{(\cmark) denotes ``high security level'' and (\xmark) denotes ``low security level.''}
	\end{tabular}
\end{table}

\begin{figure*}
	\centering
	\begin{tabular}{cccccc}
		& LE~\cite{n1} & PE~\cite{m4} & EtC~\cite{m3} &  ELE~\cite{n2} & Proposed \\ \\
		Encrypted &
		\begin{minipage}{2.0cm}
			\centering
			\includegraphics[width=2.0cm]{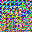}
		\end{minipage} &
		\begin{minipage}{2.0cm}
			\centering
			\includegraphics[width=2.0cm]{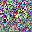}
		\end{minipage} &
		\begin{minipage}{2.0cm}
			\centering
			\includegraphics[width=2.0cm]{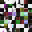}
		\end{minipage} &
		\begin{minipage}{2.0cm}
			\centering
			\includegraphics[width=2.0cm]{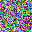}
		\end{minipage} &
	    \begin{minipage}{2.0cm}
	    	\centering
	    	\includegraphics[width=2.0cm]{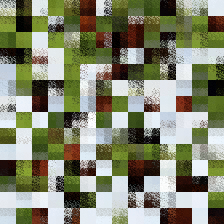}
	    \end{minipage}\\
       & 0.006 & 0.001 & 0.061 & 0.001& 0.123\\[5pt]
		FR-Attack~\cite{m5} &
		\begin{minipage}{2.0cm}
			\centering
			\includegraphics[width=2.0cm]{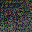}
		\end{minipage} &
		\begin{minipage}{2.0cm}
			\centering
			\includegraphics[width=2.0cm]{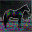}
		\end{minipage} &
		\begin{minipage}{2.0cm}
			\centering
			\includegraphics[width=2.0cm]{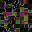}
		\end{minipage} &
		\begin{minipage}{2.0cm}
			\centering
			\includegraphics[width=2.0cm]{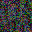}
		\end{minipage} &
	    \begin{minipage}{2.0cm}
	    	\centering
	    	\includegraphics[width=2.0cm]{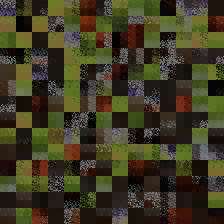}
	    \end{minipage}\\
		& 0.017 & 0.010 & 0.001 & 0.001& 0.035\\[5pt]
		GAN-Attack~\cite{n4}&
		\begin{minipage}{2.0cm}
			\centering
			\includegraphics[width=2.0cm]{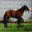}
		\end{minipage} &
		\begin{minipage}{2.0cm}
			\centering
			\includegraphics[width=2.0cm]{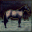}
		\end{minipage} &
		\begin{minipage}{2.0cm}
			\centering
			\includegraphics[width=2.0cm]{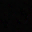}
		\end{minipage} &
		\begin{minipage}{2.0cm}
			\centering
			\includegraphics[width=2.0cm]{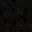}
		\end{minipage} &
	    \begin{minipage}{2.0cm}
	    	\centering
	    	\includegraphics[width=2.0cm]{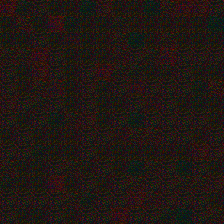}
	    \end{minipage} \\
		& 0.774 & 0.031 & 0.021 & 0.010&0.043\\[5pt]
		ITN-Attack~\cite{n5}  &
		\begin{minipage}{2.0cm}
			\centering
			\includegraphics[width=2.0cm]{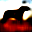}
		\end{minipage} &
		\begin{minipage}{2.0cm}
			\centering
			\includegraphics[width=2.0cm]{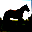}
		\end{minipage} &
		\begin{minipage}{2.0cm}
			\centering
			\includegraphics[width=2.0cm]{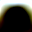}
		\end{minipage} &
		\begin{minipage}{2.0cm}
			\centering
			\includegraphics[width=2.0cm]{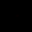}
		\end{minipage}&
		\begin{minipage}{2.0cm}
		\centering
		\includegraphics[width=2.0cm]{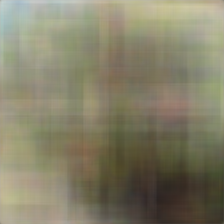}
	    \end{minipage}\\
		& 0.407 & 0.529 & 0.001 & 0.086&0.117\\
		Plain & \multicolumn{5}{c}{
			\begin{minipage}{2.0cm}
				\centering
				\includegraphics[width=2.0cm]{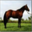}
			\end{minipage}
		}
	\end{tabular}
	\caption{Examples of images restored from encrypted ones. SSIM values are marked at bottom of  images.\label{fig:attack}}
	\label{fig:sec-eval}
\end{figure*}

\section{Experiments}
To verify the effectiveness of the proposed method, we ran a number of image classification experiments.

\subsection{Experiment Conditions}
We conducted image classification experiments on two datasets: CIFAR-10~\cite{m6} and CIFAR-100~\cite{m6} with a batch size of 512.
Both CIFAR-10 (with 10 classes) and CIFAR-100 (with 100 classes) consist of 60,000 color images (dimension of $ 3 \times 32  \times 32$), where 50,000 images are for training and 10,000 for testing.
For CIFAR-10, each class contains 6,000 images, and there are 600 images for each class in CIFAR-100.
Images in both datasets were resized to $3 \times 224 \times 224$ before applying the proposed encryption algorithm, where the block-size was $ 16\times 16$.

We used a PyTorch~\cite{p1} implementation of ViT and fine-tuned the ViT-B\_16 model which was pretrained with the ImageNet21k dataset. We used the training settings from~\cite{m2} except the learning rate.
The parameters of the stochastic gradient descent (SGD) optimizer we used were: a momentum of $0.9$, a weight decay of $0.0005$, and a learning rate value of $0.03$ for plain images and $0.3$ for encrypted ones.

\subsection{Classification Accuracy}
Table~\ref{tab:results} shows the image classification performance of the proposed method compared with the conventional methods: LE~\cite{n1}, EtC~\cite{m3}, ELE~\cite{n2}, and PE~\cite{m4}.\@
The proposed method utilized ViT denoted as ViT-B\_16, and the other methods used ShakeDrop ~\cite{m8} or ResNet-18~\cite{n7,n8}.
From the table, in any of the cases, the proposed method with ViT-B\_16 achieved the highest accuracy for both plain and encrypted images for both datasets.
Therefore, it outperformed the state-of-the-art image encryption methods for privacy-preserving image classification in terms of image classification accuracy.

\subsection{Robustness Against Various Attacks}
We performed the FR-Attack~\cite{m5}, GAN-Attack~\cite{n4}, and ITN-Attack~\cite{n5} to evaluate the robustness of the proposed encryption method on the CIFAR-10 dataset.
The robustness of LE~\cite{n1}, EtC~\cite{m3}, ELE~\cite{n2}, and PE~\cite{m4} was tested under the same conditions as in the original papers, where the image size was $ 32\times 32$ and the block-size was $ 4 \times 4$.
We used the same settings as in~\cite{n4,n5} except for some necessary changes to make these attack methods fit to the image size of $ 224\times 224$ used for the proposed method.

Figure~\ref{fig:attack} shows images restored by using the three attacks.
Structural similarity index measure (SSIM) values~\cite{m9} are marked at the bottom of the restored images to illustrate the structural similarity between a restored image and a plain one.
A larger value means a higher structural similarity between the two images.
The results from Fig.~\ref{fig:attack} indicate that the encrypted images with the proposed method did not have personally identifiable visual information in the plain images even after the attacks.
In addition, we also confirmed that the restored images for other test images in the test set followed a similar trend as in Fig.~\ref{fig:attack}.
Therefore, the proposed method was robust against such attacks.
In contrast, for the state-of-the-art encryption methods,
some visual information was restored for encrypted images with LE~\cite{n1} and PE~\cite{m4} as shown in Fig.~\ref{fig:attack}.
Although EtC~\cite{m3} and ELE~\cite{n2} were robust against such attacks, the classification accuracy dropped significantly (Table~\ref{tab:results}).

\section{Conclusion}
In this paper, we proposed the combined use of a novel learnable encryption algorithm and ViT for privacy-preserving image classification. The proposed method enables us not only to use visually protected images but to also maintain a high classification accuracy. In experiments, the proposed method was demonstrated to outperform state-of-the-art methods with perceptually encrypted images in terms of classification accuracy and robustness against various attacks.

\section{Acknowledgement}
This study was partially supported by JSPS KAKENHI (Grant Number JP21H01327), JST CREST (Grant Number JPMJCR20D3), and Support Center for Advanced Telecommunications Technology Research, Foundation (SCAT).

\bibliographystyle{IEEEtran}
\bibliography{IEEEabrv,refs}
\end{document}